\documentclass{article}

\usepackage{microtype}
\usepackage{graphicx}
\usepackage{subfigure}
\usepackage{booktabs} 

\usepackage{hyperref}

\usepackage[accepted]{icml2023}

\usepackage{amsmath}
\usepackage{amssymb}
\usepackage{mathtools}
\usepackage{amsthm}

\usepackage[capitalize,noabbrev]{cleveref}

\newcommand{\RN}[1]{%
  \textup{\uppercase\expandafter{\romannumeral#1}}%
}

\theoremstyle{plain}

\theoremstyle{definition}

\theoremstyle{remark}

\usepackage[textsize=tiny]{todonotes}

\icmltitlerunning{Social learning spontaneously emerges by searching optimal heuristics with deep reinforcement learning}

\begin{document}

\twocolumn[
\icmltitle{Social learning spontaneously emerges by searching optimal heuristics \\ with deep reinforcement learning}




\begin{icmlauthorlist}

\icmlauthor{Seungwoong Ha}{kaist}
\icmlauthor{Hawoong Jeong}{kaist,ccs}

\end{icmlauthorlist}

\icmlaffiliation{kaist}{Department of Physics, Korea Advanced Institute of Science and Technology, Daejeon 34141, South Korea}
\icmlaffiliation{ccs}{Center for Complex Systems, Korea Advanced Institute of Science and Technology, Daejeon 34141, South Korea}

\icmlcorrespondingauthor{Hawoong Jeong}{hjeong@kaist.edu}

\icmlkeywords{Social Learning, Reinforcement learning, Heuristics}

\vskip 0.3in
]



\printAffiliationsAndNotice{\icmlEqualContribution} 

\begin{abstract}
    How have individuals of social animals in nature evolved to learn from each other, and what would be the optimal strategy for such learning in a specific environment? Here, we address both problems by employing a deep reinforcement learning model to optimize the social learning strategies (SLSs) of agents in a cooperative game in a multi-dimensional landscape. Throughout the training for maximizing the overall payoff, we find that the agent \textit{spontaneously learns} various concepts of social learning, such as copying, focusing on frequent and well-performing neighbors, self-comparison, long-term cooperation between agents, and the importance of balancing between individual and social learning, without any explicit guidance or prior knowledge about the system. The SLS from a fully trained agent \textit{outperforms} all of the traditional, baseline SLSs in terms of mean payoff. We demonstrate the superior performance of the reinforcement learning agent in various environments, including temporally changing environments and real social networks, which also verifies the adaptability of our framework to different social settings.
\end{abstract}

\section{Introduction}

Learning is one of the most salient properties that emerge from flourishing species in nature. Particularly, learning from other members in a group generally leads to a collective success by exploiting verified information \cite{feldman1984cultural, feldman1996individual, kendal2018social}, which distinguishes social learning from asocial, individual learning where information directly comes from exploration through the environment. While social learning is intuitively beneficial at first sight, research over the past several decades has consistently proven that naive social imitation is not inherently adaptive and often fails to achieve good group-level performance \cite{boyd1988culture, laland2004social, hashimoto2010new, rendell2011cognitive, mason2012collaborative, barkoczi2016social, todd2020interaction}. Instead, current theory suggests that, to properly determine how to learn from others one should employ a selective heuristics called a social learning strategy (SLS)  \cite{laland2004social}, which governs the internal rules for choosing the proper time, subject, and methods to engage both social and individual learning. SLSs significantly contribute to building social norms and driving cultural evolution in society, and thus the understanding of SLSs provides fruitful insight to policymakers and group leaders \cite{toyokawa2019social, almaatouq2020adaptive}. Throughout the history of research into SLSs, two fundamental questions have yet to be fully answered: how it naturally emerged for social beings in nature, and what is the optimal strategy for the given environment.

In this work, we employ a modern computational model to tackle both of the central questions---namely, regarding the natural emergence of social learning and finding the optimal strategy. By constructing a reinforcement learning (RL) framework with a neural network tailored to SLSs, we train a model-free agent to search for the multi-dimensional policy that yields the maximum average payoff during its social evolution. We show that social learning in a cooperative game can naturally emerge through RL from a simple reward, without any selective pressure or explicit knowledge of the information the agent receives from interactions. The most intriguing point is that the agent progressively discovers significant notions of social learning throughout the training, including the concept of copying other solutions based on their frequency or payoff, stochastic copying, individual learning, self-comparison, and even the delicate interplay between exploration and exploitation, which yields a long-term cooperation between agents in a non-greedy manner. The optimized SLS from the trained agent outperforms all of the baseline SLSs in various environmental settings, including real social networks.

\begin{figure*}
    \centering
    \includegraphics[width=0.9\linewidth]{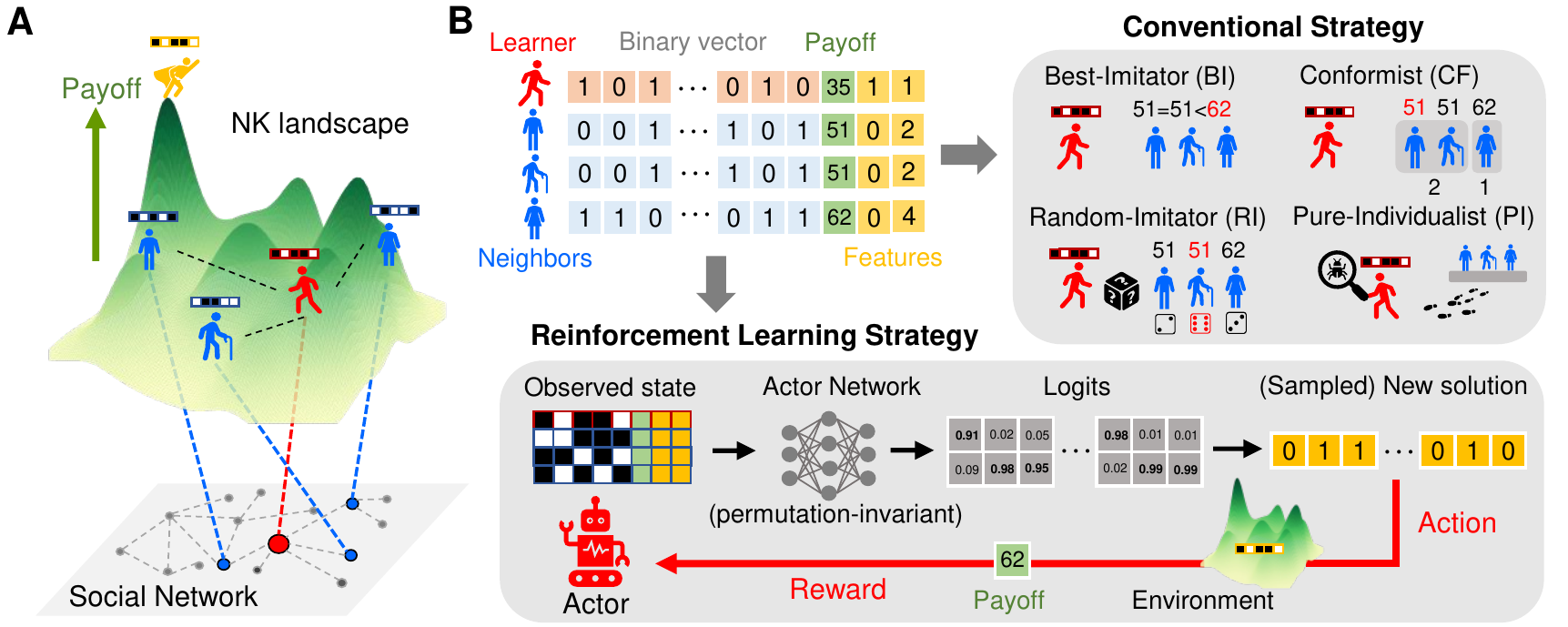}
    \caption{(A) NK model on a social network. At every time step, each person randomly observes a certain number of people among their neighbors and applies a social learning strategy (SLS) to maximize their individual time-average payoff. The solution is represented as an $N$-dimensional binary vector, where its payoff is given by the NK landscape. (B) Available information and various mechanisms for SLSs. The conventional strategies include frequently observed and proposed heuristics, such as best imitator and conformist strategies, while we present a reinforcement learning approach to find the optimal SLS for the given environment. The agent tries to maximize the time-average payoff it receives from the NK landscape as a result of its action of producing probabilities for a new solution. State correction after the sampled solution is omitted for visualization.}\label{fig:1}
  \end{figure*}

\section{Related studies}

Many studies attribute the emergence and evolution of social learning to natural selection \cite{boyd1988culture,laland2004social, macy1996natural, perreault2012bayesian}, while mostly explaining the origin of social learning in a retrospective manner. Some evidence has shown that reinforcement learning could lead to the emergence of social learning \cite{dawson2013learning, lindstrom2015mechanisms}, but a general framework for inducing complex SLSs is still lacking. On the other hand, numerous attempts to search for the best SLS have faced another set of problems. Previous studies mainly focused on performing a comparison or organizing a tournament between a given set of heuristics \cite{rendell2010copy,csaszar2010much,mason2012collaborative, fontanari2015exploring, barkoczi2016social, barkoczi2016collective}, which have been either reported from empirical societies or proposed by insights from social observations without a rigorous theoretical basis or optimization scheme. Although a number of computational models and theoretical approaches attempted to formalize the long-term behavior of SLSs \cite{aoki2005emergence,lopes2009computational, aoki2010evolution, giraldeau2018social}, optimizing general SLSs even in a simple environment is an extremely challenging task due to the inherent multifaceted complexity, such as from an exponentially large search space, dependence on interaction networks, non-differentiable payoff, and stochasticity.

To search heuristics systematically without brute force, metaheuristics such as genetic algorithms \cite{sivanandam2008genetic} and simulated annealing \cite{van1987simulated} are widely used. Recently, fueled by the rapid advances in machine learning, many researchers have started to employ RL to seek heuristics \cite{runarsson2011learning, bianchi2015transferring, fiderer2021neural, yonetani2021path, yaman2021meta}. In particular, RL has shown its strength in constructing computational models of the spatiotemporal dynamics of human cooperation and anthropological problems \cite{vinitsky2021learning, ndousse2021emergent, mckee2021deep, koster2022spurious}. In this paper, we aim to adopt the similar spirit to these and employ the full potential of RL framework to find the optimal SLS in the given settings.

\begin{figure*}
    \centering
    \includegraphics[width=1\linewidth]{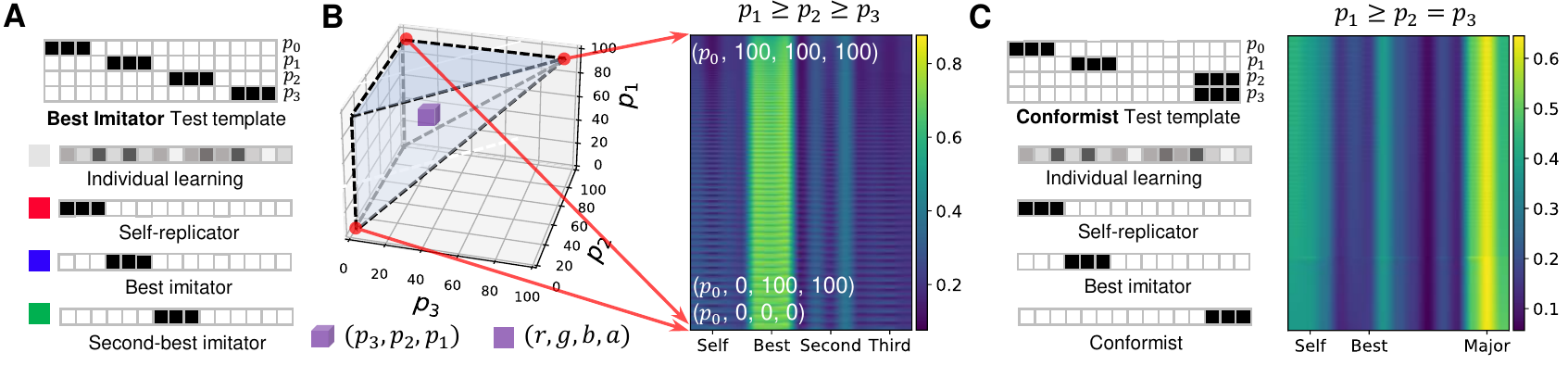}
    \caption{(A) BI test template solution vectors and representative strategies for the face colors of the voxels. (B) Three-dimensional (3D) strategy diagram (left) and two-dimensional (2D) output diagram (right) for the BI test. The model results from the test template with score $p_1\leq p_2\leq p_3$ are shown. In the 3D diagram, each voxel's location represents its neighbor's score $(p_1, p_2, p_3)$ and its face color indicates the distances between the output probability and identified strategies. The 2D output diagram directly shows the output probability of producing $1$ for each $N$ dimension. For visualization, in what follows, the voxels in the 3D strategy diagrams are drawn by taking every 5th coordinate value in each dimension. (C) CF test template solution vectors and representative strategies (left) and 2D output diagram (right) employing the test template with $p_1\leq p_2=p_3$ to accuratley detect the true CF strategy.}\label{fig:2}
  \end{figure*}

\section{Methods}

  Here, we model the problem of social learning by considering a group of individuals that iteratively search for better solutions in a rugged, high-dimensional landscape \cite{csaszar2010much, hashimoto2010new, fontanari2015exploring, barkoczi2016social, giannoccaro2018team, todd2020interaction}, where our goal is to find the optimal heuristic for individuals that yields the maximum average payoff when shared with its group (Fig. \ref{fig:1}A). In this paper, the rugged landscape takes the form of an NK model \cite{kauffman1987towards}, see Appendix for further details.

  We formulate SLSs as intrinsic algorithms for an individual in a group who receives information from their neighbors and yields the probability of their future solution for every time step. The collective information comprises solutions and payoffs as well as additional features such as rankings and frequencies, but these are provided without any indication; the agent is initially clueless about which part of the information is a payoff or solution. This stochastic formulation can encompass both social learning and individual learning in a unified framework in multi-dimensional settings (Fig. \ref{fig:1}B). One notable point is that the algorithms should be invariant to permutations of the neighbors' information, since generally there is no specific order of neighbors. We designed the neural architecture to properly handle this characteristic by adopting a permutation-invariant encoder for the policy network (see Appendix for further details).

As a baseline, we consider the following strategies from previous literature \cite{laland2004social, barkoczi2016social,barkoczi2016collective}. Best imitator (BI) always copies the solution of the best-performing neighbor, conformist (CF) always copies the most frequent (or major) solution among the neighbors, random imitator (RI) chooses random neighbors to copy, and pure individualist (PI) does not engage in any form of social learning (Fig. \ref{fig:1}B). For SLSs with individual learning, single-bit flipping (-I), probabilistic flipping (-P), or random flipping (-R) are applied to the current solution (see Appendix for further details).

Since the strategies are formulated as high-dimensional functions, understanding and visualizing the functional meaning of a trained neural network is not a simple task. Here, similar to controlled experiments in psychology, we inspect the strategy of the trained RL agent by isolating it from the network and observe the solution yielded by the policy network using a given test template. We test the similarity of the given model output to two representative SLSs, i.e., BI and CF strategies. The BI test template (Fig. \ref{fig:2}A) consists of a series of fixed solution vectors and tunable payoffs, $0 \leq p_0 \leq p_{\text{max}}=100$ for the learner itself and $0 \leq p_3 \leq p_2 \leq p_1 \leq p_{\text{max}}$ for the neighbors in decreasing order of payoff. Since the proper heuristics should only depend on the payoff and not on the form of the solution vector itself, we can investigate the nature of the SLSs by changing the payoffs $(p_0, p_1, p_2, p_3)$ and observing the output probabilities. For instance, we can expect that the BI-I strategy will imitate the solution of $p_1$ if $p_1 > p_0$ and perform individual learning otherwise, regardless of $p_2$ and $p_3$. The CF test template (Fig. \ref{fig:2}C) is constructed in a similar manner, but in this case, two of the solutions are the same with a low payoff ($p_1\leq p_2=p_3$) to precisely discern whether the agent follows the major solution even if it is worse than the other solution. For both tests, we draw a two-dimensional (2D) output diagram from $(p_0, 0, 0, 0)$ to $(p_0, 100, 100, 100)$ that satisfies the respective payoff conditions ($176,581$ pairs for the BI test and $5,050$ pairs for the CF test). 

For visualization, in Fig. \ref{fig:2}B we depict the response of these SLSs as a three-dimensional (3D) voxel plot for a fixed $p_0$, where each voxel is located at $(p_3, p_2, p_1)$ with a face color $(r, g, b, a)$. Each RGB color component represents the distance between the given strategy and the specific solution, as visualized in Fig. \ref{fig:2}A, and the opacity $a$ depends on the minimum distances among all of the solutions (see Appendix for details). With this color scheme, the voxel shows PI-R as light gray (low opacity, hence not close to any of the given solutions), PI-I as translucent red (close to the self solution but with some randomness), and BI as vivid blue (identical to the best solution). This type of 3D strategy diagram, along with the 2D output diagram, enables us to investigate the qualitative characteristics of the agent's multi-dimensional strategy that could not be easily comprehended otherwise.

\begin{figure*}[!ht]
  \centering
  \includegraphics[width=0.9\linewidth]{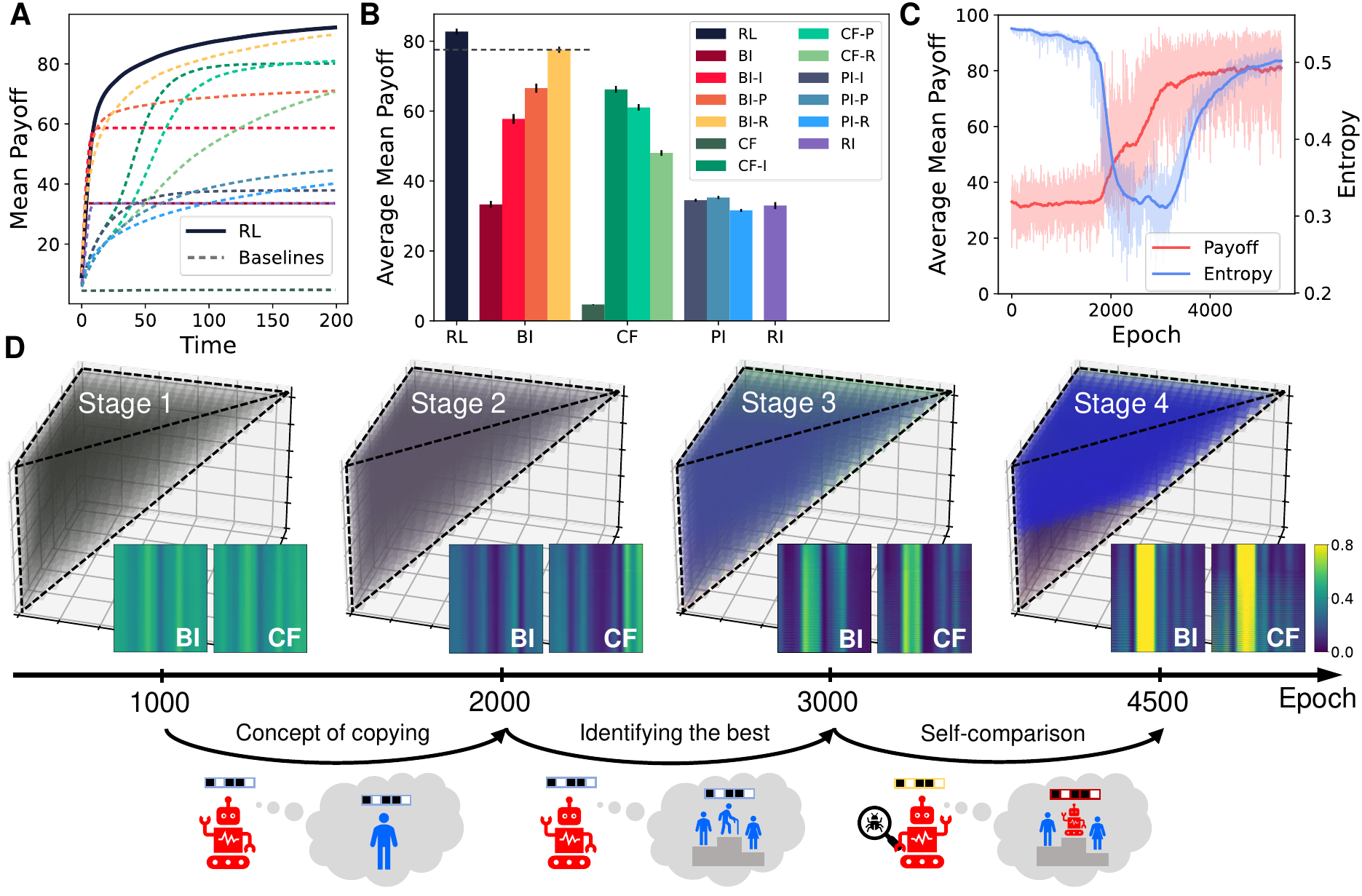}
  \caption{(A) Mean payoff and (B) average mean payoff over time of the SLS from reinforcement learning (RL) and various baseline SLSs (BI, BI-I, BI-P, BI-R, CF, CF-I, CF-P, CF-R, PI-I, PI-P, PI-R, RI). Here, the RL agent ($83.03$) surpasses the best-performing baseline, BI-R ($77.54$), as indicated with the dotted line in (B). Error bars show $\pm$5 standard error of the mean. (C) Average mean payoff and entropy of the model output during $5,700$ training epochs. The bold lines show the exponential moving average with a smoothing factor of $0.99$. (D) 3D strategy diagrams and 2D output diagrams for the BI test and the CF test from the model with training epochs of $1,000$, $2,000$, $3,000$, and $4,500$, each representing $4$ distinct stages of learning dynamics. Below the plots, we draw conceptual diagrams of the important lessons that the agent learned as the agent passes through to the next stage. The model is trained on a fully connected network of $100$ agents with NK$(15, 7)$ environments.}\label{fig:3}
\end{figure*}

\section{Results}

\subsection{Default environment}

In each epoch of the model training, $100$ agents with randomly assigned initial states perform the same SLS with $3$ randomly selected neighbors for $L=200$ time steps, and the reward for each agent is given as the payoff from the environment  according to each individual's new state. Even though this effectively trains a single model with a group of self-copied agents, we grant rewards \textit{individually}, and thus the model is optimized to maximize the expected payoff of each individual agent. Accordingly, from each individual's perspective, the solutions of all other agents and payoffs are regarded as surroundings, not a subject of optimization. We find that the model significantly struggles when a group-averaged reward is provided instead of an individual reward (see Supplementary Information for the result).

We set our default environment as NK$(15,7)$ on the fully connected network, a similar setting as \cite{barkoczi2016social}, and train the agent by providing a new random landscape every epoch. This learning scheme is critical for guiding the agent to learn general heuristics rather than a single solution, as we find that the model output converges to a single optimal strategy when only a small number of fixed landscapes are given (see Supplementary Information for the results when $1$ and $10$ fixed environments are given to the agent). In this experiment, we train the agent for $5,700$ epochs, and the final model is used to measure the performance. Results reported are averaged across $5,000$ repetitions.

First, we compare the performance of the SLS from RL with those of $12$ baseline SLSs by measuring the mean payoff from all agents of multiple trials and multiple initial landscapes, as shown in Fig. \ref{fig:3}A and \ref{fig:3}B (see Appendix for details). Here, the average mean payoff corresponds to the area under the curves in Fig. \ref{fig:3}A divided by the total time $L=200$, while the final mean payoff corresponds to the mean payoff value at the final time, $200$. The result clearly shows the dominant performance of the SLS from RL, exceeding the average payoff of all baselines by a noticeable margin.

Obviously, this overwhelming performance of the trained agent is not inherent from the beginning; the agent initially performed poorly and gradually improved via learning, as visible in Fig. \ref{fig:3}C. This strongly implies that the model somehow acquired the ability of social learning during the learning process. Another noteworthy point is that the agent's average mean payoff constantly increased while the entropy of the output distribution showed non-monotonic behavior during the training. The entropy of an output distribution directly assesses how \textit{confident} each dimension of a solution is; for binary cases, low entropy indicates that the probability of producing $1$ is close to either $0$ or $1$, rather than being indecisive and having a probability of $0.5$. Hence, the non-monotonic behavior of entropy indicates that the model converged into a certain solution, but the convergence was abandoned spontaneously and the strategy became more random again. To further investigate these peculiar learning dynamics, we plot a 3D strategy diagram and 2D output diagrams for both tests from the model after $1,000$, $2,000$, $3,000$, and $4,500$ training epochs, when $p_0 = 50$ (Fig. \ref{fig:3}D). Critically, we find that the agent passes through four unique sub-optimal strategies during the learning process before it reaches the final strategy.

The model starts from a totally random strategy that produces a probability of near $0.5$ regardless of the input, similar to the PI-R strategy as expected. This is the stage where the agent does not utilize the information from its neighbors, i.e., it has no concept of social learning at all. The emergence of referencing others appears after gaining extensive experience from iterated games, in this case, around $1,500$ epochs (Stage 2 in Fig. \ref{fig:3}D). Surprisingly, the first thing that the agent learns is to copy the major solution, similar to the CF strategy, which is accompanied by a drastic decline in entropy. Although the concept of 'copying the major solution' is generally not the best-performing, as reflected in the CF-based strategies in Fig. \ref{fig:3}B, this concept can be easily captured by an agent since we provide the frequency of each solution as a feature. We speculate that passing through this sub-optimal strategy facilitates faster learning by providing additional options to copy and helping the agent learn the notion of copying much faster, similar to a recently reported phenomenon in artificial agents for social learning \cite{koster2022spurious}. This is further supported by a delay in the learning process when the frequency feature is not provided (see Supplementary Information for the result where no solution frequency is given). In Stage 2, the agent finds a connection between observed information and its behavior, which can be likened to the acquirement of neurophysiological circuits such as mirror neurons \cite{dickerson2017role, olsson2020neural}.

After the concept of copying is well understood by the agent, the subject that is copied quickly transfers from the major solution to the best solution around $2,500$ epochs (Stage 3 in Fig. \ref{fig:3}D). We find that the agent needs a far greater number of training epochs to reach the final strategy if it has to learn the payoff ranking by itself (see Supplementary Information for the result where the payoff ranking is not provided). After this subject switching, the agent gets confident with the strategy of 'copying the best solution', as shown in vivid blue in the 3D diagrams in the figure. The agent in this stage shows a similar strategy to the BI strategy with a small chance of flipping.

\begin{figure}
    \centering
    \includegraphics[width=\linewidth]{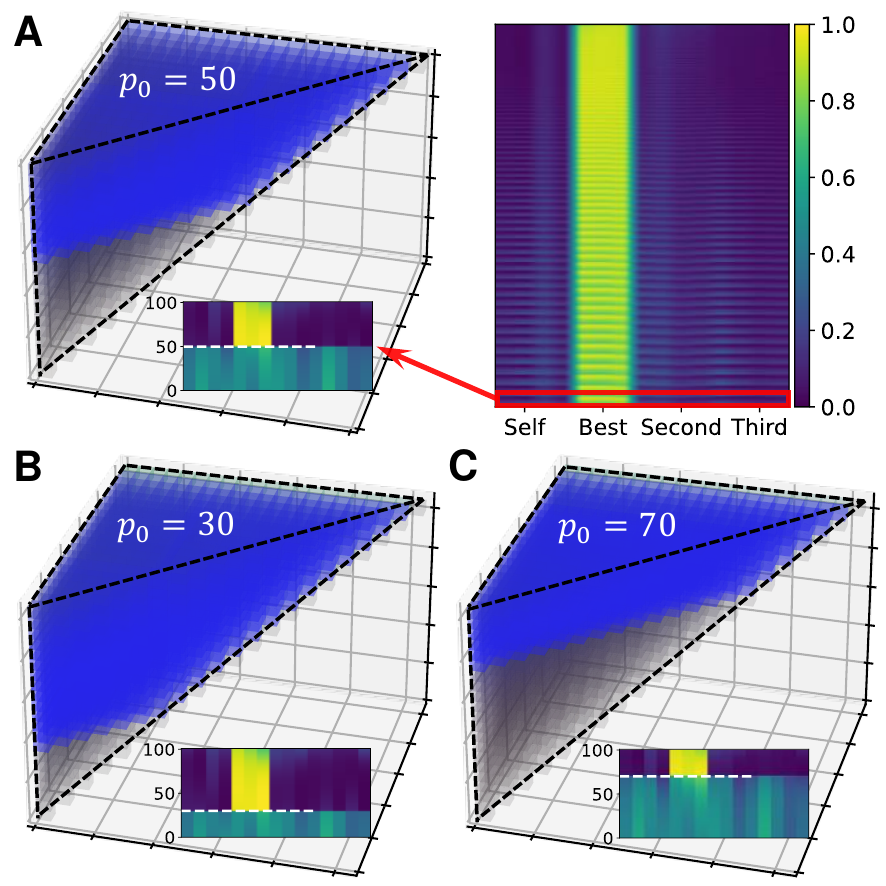}
    \caption{(A) A 3D strategy diagram and 2D output diagram from the final model of the default environment with $p_0=50$. The zoomed-in inset shows the output probability from $p_2=p_3=0$ and $0\leq p_1 \leq 100$ to visualize the clear strategic boundary at $p_1=50$, highlighted with a white dotted line. (B, C) 3D strategy diagrams and corresponding insets with $p_0=30$ (B) and $p_0=70$ (C). The model is trained on a fully connected network of $100$ agents with NK$(15, 7)$ environments.}\label{fig:4}
  \end{figure}
  
The last stage of learning starts with a rapid increase in entropy after $3,000$ epochs (Stage 4 in Fig. \ref{fig:3}D). This does not indicate that the model backslides to the very beginning, however; the agent clearly maintains the lessons from former experiences in some conditions while intentionally forgetting them in other conditions according to a specific threshold. The appearance of the translucent lower region of the 3D diagram and greenish stripes in the 2D output diagram from the model of $4,500$ epochs demonstrates such a transition visually. It turns out that the agent at this stage begins to compare the payoff of the best solution ($p_1$) to its own payoff ($p_0$) and begins to choose to employ a different strategy based on the comparison. By observing the final model's 3D strategy diagram and 2D output diagram for different $p_0$ (Fig. \ref{fig:4}), it is evident that the agent performs a random search when $p_0 \geq p_1$ and imitates the best solution when $p_0 < p_1$. In this final stage, the agent exhibits every key aspect of the BI-R strategy, the best-performing baseline SLS in the environment.

This remarkable transition is one of the key findings of our work. Note that from each individual's perspective, performing a totally random search gives a minuscule probability of finding a better solution (especially if the agent already achieved a high payoff), hence it cannot be properly incentivized by an optimization scheme without long-term planning. From stage 3 where every agent blindly copies others, our RL agent \textit{realizes} that no one, including itself, could achieve a better solution if nobody provides a new solution to the population pool. This leads the agent from an information scrounger to an information producer \cite{laland2004social} by abandoning some part of selfish social learning and developing individual learning which also benefits other agents. In stage 4, the agent finds the exact division between two learning schemes, achieving the most effective form of long-term cooperation. Also, we find that the agent here deliberately chooses to perform a totally random search (similar to ``-R'' individual learning) among the various forms of individual learning it may adopt. See Supplementary Information for the movie of 3D strategy diagrams throughout the full training, which exhibits various short-lived strategical notions such as 'copying the second-best' and individual learning that adopts randomness while preserves the current solution.

Throughout this detailed analysis, we demonstrate both the variety of SLSs that can be expressed by our model and the capability of the RL framework for observing the transmission of behavior by social interactions. Note that we did not incentivize any social behaviors by explicitly providing the means or assigning specific rewards; rather, our framework provides only raw information from randomly chosen neighbors without any prior knowledge. By employing a model-free computational approach with neural networks, we show that diverse and complex social learning strategies in nature can spontaneously emerge from the simplest reward with sufficient social interactions.

\begin{figure*}[!ht]
    \centering
    \includegraphics[width=0.9\linewidth]{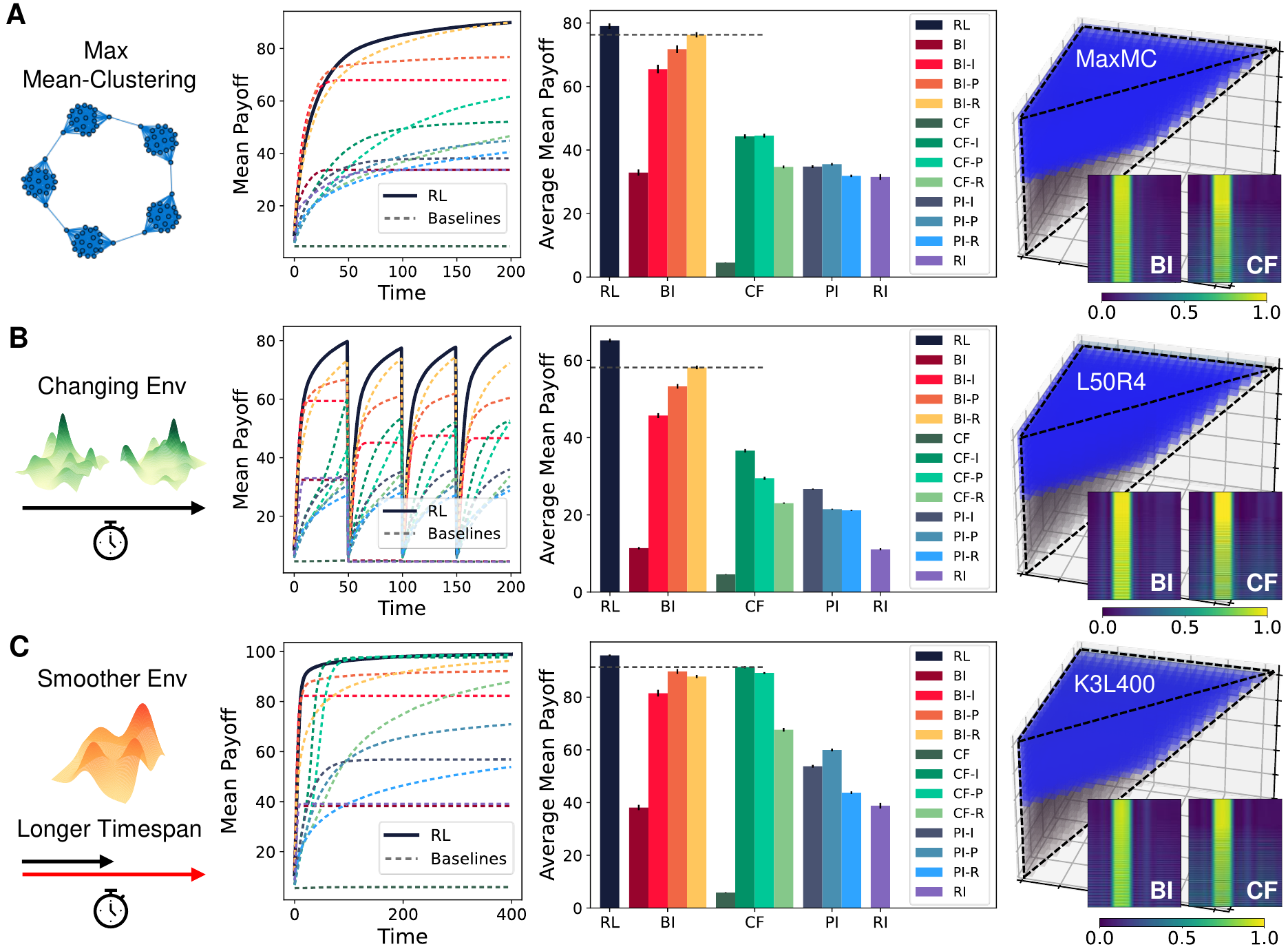}
    \caption{Trained model results from diverse environment settings compared to default. From left to right, each panel shows a conceptual diagram, mean payoff, average mean payoff over time, and 3D strategy diagram with 2D output diagrams for both BI and CF tests. (A) Results from a different underlying network. Instead of a complete network, agents are now connected in a modular network that is optimized to maximize the mean clustering coefficient (MaxMC). (B) Results from a temporally changing environment. Every $50$ time steps (hence $4$ times among $200$ time steps), the underlying NK landscape is randomly reassigned, and all of the scores are newly calculated based on the new landscape (L50R4). (C) Results from a smoother environment with a longer time span. The landscape becomes smoother with $K=3$, and the total time steps are increased to $400$ (K3L400). }\label{fig:5}
  \end{figure*}

\subsection{Various different environments}

One of the major advantages of the computational approach for social learning is that we can freely alter the characteristics of the given environment that reflects the various learning dynamics observed in the real world. In Fig. \ref{fig:5}, we present the performance and strategies of the final models trained with three different environmental settings. 

First, we change the network structure between agents to alter the speed of information spread \cite{derex2016partial, smolla2019cultural} (Fig. \ref{fig:5}A). The network we use, the \textit{Max mean clustering} (MaxMC) network, is directly adopted from \cite{barkoczi2016social}. It consists of the same $100$ agents as the fully connected default network but has a lower degree ($19$ links for each node) and is optimized to maximize the mean clustering coefficient (see Appendix for details). We choose this network as an extreme case of slow information spread, as this network was employed by the original authors to demonstrate the effect of a high network diameter on the performance of SLSs. Next, we reset the landscape every $50$ time step to simulate a temporally changing environment \cite{aoki2005emergence, hashimoto2010new, giannoccaro2018team, toyokawa2019social} (Fig. \ref{fig:5}B). Since each $R=4$ landscape lasts $L=50$ time steps, we call this experiment L50R4. Lastly, we smoothen the landscape by setting $K=3$ and lengthen the time span to twice the default game, $L=400$ (Fig. \ref{fig:5}C). This specific setting (K3L400) is deliberately chosen to let a CF-based SLS become the best-performing baseline (in this case, CF-I), while also demonstrating the variety of environmental settings that can be controlled.

From the results in Fig. \ref{fig:5}, all three agents exhibit strategies similar to BI-R, which appear to be outstanding in every condition. We find that in some cases, the agent may stay longer at a certain stage compared to the default environment, but eventually, the model transits from such sub-optimal strategies and converges to a final strategy. Even in the K3L400 environment, the agent's final strategy does not reference the major solution but still exceeds the best-performing baseline, CF-I. This result suggests that the 'copying the best solution' strategy is indeed powerful, especially when it is accompanied by enough randomness from individual learning; both characteristics are successfully discovered by our RL framework. We also apply our framework to other environments, including a much more rugged landscape ($K=11$) and $53$ different real social networks from \cite{ghasemian2020stacking}, the agents of which again show superior performances compared to the baselines (See Supplementary Information for results).

\begin{figure*}
  \centering
  \includegraphics[width=\linewidth]{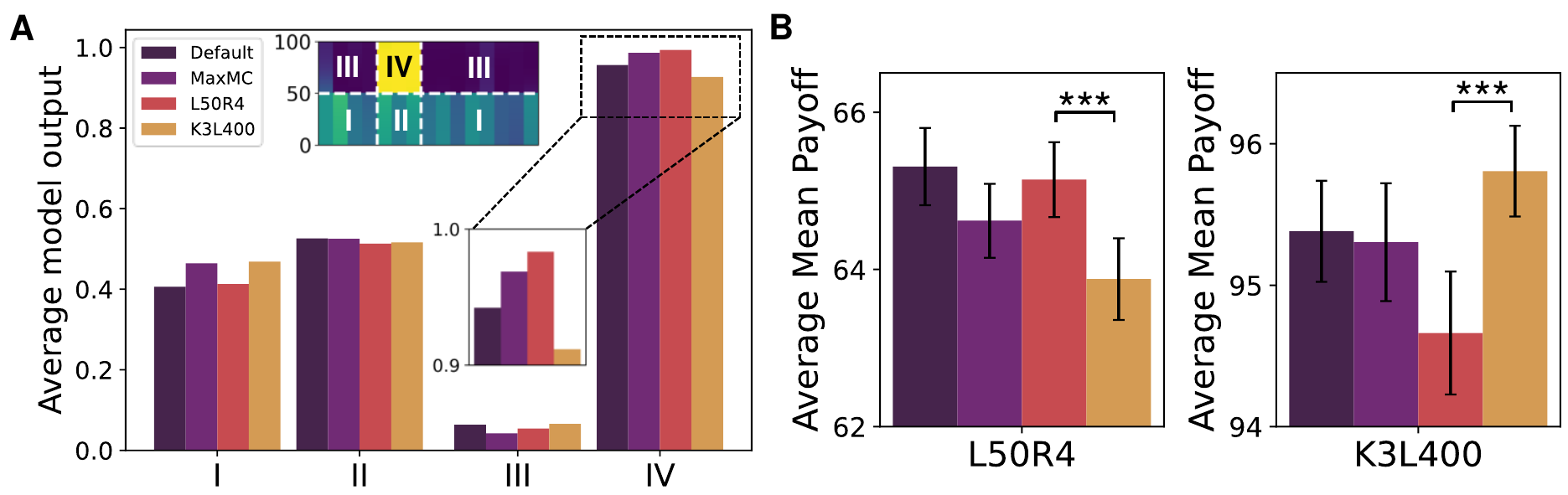}
  \caption{(A) The average output probability of four regions from the BI test template with $p_2=p_3=0$, calculated for $4$ models (Default, MaxMC, L50R4, K3L400) trained in different environment settings. The upper diagram is a portion of the 2D output diagram when $p_2=p_3=0$ and $0\leq p_1 \leq 100$ showing the division of the four regions. The output is averaged for every pair of $0\leq p_0\leq  p_1\leq 100$. Here, regions \RN{1} and \RN{2} indicate the dimensions of the non-best and the best solution when $p_0\geq p_1$, respectively, while regions \RN{3} and \RN{4} indicate the dimensions of the non-best and the best solution when $p_0 < p_1$, respectively. The zoomed-in inset emphasizes the difference of degree of copying between each model by magnifying the result from region \RN{4}, which has values of $0.9562$ (default), $0.9863$ (MaxMC), $0.9941$ (L50R4), and $0.9269$ (K3L400). (B) Average mean payoff of the $4$ models evaluated in L50R4 (left) and K3L400 (right) environments. ***$p<0.001$.}\label{fig:6}
\end{figure*}

\subsection{Comparison between specialized agents from different environments}

With a careful inspection, one may notice that the 2D output diagrams of the three models in Fig. \ref{fig:5} are not perfectly identical. Even though we described all three agent's strategies as ``similar to BI-R'', they are in fact not the same. We observe that there are critical differences among these strategies, and that each model adapted differently to maximize the reward in their given environment.

The difference between strategies becomes clear when we plot the average model output (probability of producing $1$) of the agents from default, MaxMC, L50R4, and K3L400 environments, by applying the BI test template with $p_2=p_3=0$ and $0\leq p_0, p_1\leq 100$ (Fig. \ref{fig:6}A). For every $0\leq p_0\leq 100$, we compute the model output with $p_0\leq p_1\leq 100$ and separately measure the average in four different regions; regions \RN{1} and \RN{2} correspond to the dimensions of the non-best and best solutions when $p_0\geq p_1$, while regions \RN{3} and {4} correspond to the dimensions of the non-best and best solutions when $p_0<p_1$, respectively. For example, the average value of the BI-R strategy with this division would be $0.5$ in regions \RN{1} and \RN{2} (due to random individual learning), $0.0$ for region \RN{3}, and $1.0$ for region \RN{4}. Focusing on region \RN{3}, we see that none of the trained agents show the exact value of $1.0$, which implies that all of them are copying the best solution with some chance of flipping. In the real world, this kind of stochastic copying can occur due to the intrinsic noise of the copying mechanism or some level of persistence in following an individual's own traits. 

Among the agents, the agent from the L50R4 environment has the highest possibility of copying ($0.9941$) while the one from K3L400 has the lowest ($0.9269$). To find out whether this difference is evidence of adaptation to their different environments or not, we evaluate the average mean payoff of the models on these two opposite ends of the spectrum, namely, the L50R4 and K3L400 environments (Fig. \ref{fig:6}B). We find that L50R4 and K3L400 are the best-performing model in their own environment, but show significantly low performance in the opposite environment. This strongly testifies to the fact that the difference in copying probability is a result of a proper adaptation to the environment. We speculate that the high chance of flipping in the agent from K3L400 is related to the fact that CF-based strategies, which generally involve a lower level of convergence, show better performances in the K3L400 environment. To summarize, reinforcement learning effectively guides the agent to the optimal SLS for the given environment by tuning the delicate balance between exploration and exploitation, a process which could not be achieved without an exhaustive search.

\section{Conclusion and Outlook}

Different from previous studies \cite{vinitsky2021learning, mckee2021deep, koster2022spurious}, the payoff of our work is given by a fixed landscape, not from a game between agents with a payoff matrix. Our work suggests that social learning can emerge even when explicit payoff interaction between agents is not present, which resonates with the importance of vicarious reinforcement \cite{bandura1963vicarious} in social learning theory. With enough social interaction and observable information, we show that a simple motivation of payoff maximization can lead an individual to an advanced strategy of social learning.

Since our RL framework opens a new way to explore a vast space of social heuristics, one may alter the assumptions of the present work like we changed the episode length and network structure. For example, the observable payoff could be indirect and noisy \cite{horner2010prestige, mackintosh1971analysis, van2010selective}, or the strategy may involve time-dependent memory such as the social learning of multi-agent multi-armed bandits \cite{vial2021robust, sankararaman2019social, rendell2010copy}. Also, agents in nature often perform multiple strategies at once \cite{kendal2018social, nakahashi2012adaptive, bordignon2021adaptive} and do not randomly choose their subjects from among their neighbors \cite{laland2004social}, which might need more elaborate architectures to model. By implementing a permutation-invariant neural network as a policy generator and using reinforcement learning, our framework is versatile enough to integrate a variety of intriguing social characteristics.

Still, there are several limitations to the developed framework. Clearly, our stochastic formulation and neural implementation cannot express \textit{every} possible SLS due to limitations in both modern neural networks and the formulation itself. For instance, the flipping of exactly one random bit in a solution cannot be precisely expressed with our formulation. Also, the investigation and visualization of our model mainly focused on the similarity to already known strategies, which implies that there could be some hidden behavior that is complex enough to be undetected by our test templates (see Supplementary Information for more discussion). 
Moreover, societies in nature consist of heterogeneous groups of agents, experience mating and the birth-death process \cite{smolla2019cultural}, and the acquired knowledge of SLSs is not instantly adopted by every constituent.
Extending the current work to incorporate such biological and cognitive processes would be an intriguing research direction.

To sum up, we developed a neural architecture and training regime that yields complex social learning strategies spontaneously from scratch. Our study has broad implications for social norm formation, cultural evolution \cite{vinitsky2021learning}, and the exploration/exploitation trade-off \cite{mehlhorn2015unpacking}. We highlight that the successful modeling of SLSs via reinforcement learning can provide plausible evidence for the superiority of certain SLSs observed in the real world and also the detailed dynamics of their emergence.


\section{Data availability}
Simulation code and data files are available at https://github.com/nokpil/SocialNet.

\section{Acknowledgments}
  This research was supported by the Basic Science Research Program through the National Research Foundation of Korea NRF-2022R1A2B5B02001752.

\bibliography{SocialNet}
\bibliographystyle{icml2023}

\newpage
\appendix
\onecolumn

\section{Detailed methods}

\subsection{Stochastic formulation of multi-dimensional SLSs}
In this work, we define SLSs as stochastic heuristics that aggregate information from observations and produce a probabilistic expression of a future state. More rigorously, we consider the state of the $i$-th agent at time step $t$, $\mathbf{X}_i^t$, as an $N$-dimensional vector of categorical variables. Here, a state can be regarded as a solution for a landscape or a result of an action performed by an agent. Each of the dimensions of the state, $x_{i,d}^t$, may have a different number ($s_d$) of possible actions, namely, $x_{i,d}^t \in {a_d^1, a_d^2, \cdots a_d^{s_d}} = \mathcal{A}_d$. The joint action space is then defined as $\textbf{X}_i^t = (x_{i,1}^t, x_{i,d}^t, \cdots, x_{i,N}^t) \in \mathcal{A}_1 \times \mathcal{A}_2 \times \cdots \times \mathcal{A}_{N} = {\mathcal{A}}$. Also, we denote the payoff as $p_i^t$, the additional features as $\mathbf{I}_i^t$, and the aggregated information from a single agent as $\mathbf{G}_i^t = (\mathbf{X}_i^t, p_i^t, \mathbf{I}_i^t)$. With these notations, an SLS is formulated as a stochastic transition function, $\text{Pr}(\mathbf{X}_i^{t+1} \in \mathcal{A}) = \mathcal{F}(\mathbf{G}_i^t, \mathbf{G}_{j_1}^t, \cdots, \mathbf{G}_{j_S}^t)$, where the indices $(j_1, j_2, ... j_S)$ indicate $S$ interacting neighbors.

\subsection{Task environment generation}
We employ the NK landscape \cite{kauffman1987towards}, which is multi-peaked and tunably rugged, for the task environment of the social learning strategies (SLSs). The NK landscape assigns a payoff to a binary $N$-dimensional vector by averaging the contributions of the $N$ element, where each contribution is dependent on $K - 1$ other elements that are randomly determined at the initial construction. Precisely, given payoff function $f(N_i|N_i, N_{i+1}, \dotsc, N_k)$, the total payoff $P$ is $\frac{1}{N}\sum^N_{i=1}f(N_i|N_i, N_{i+1}, \dotsc, N_k)$. We set each $f(N_i|N_i, N_{i+1}, \dotsc, N_k)$ as a random number drawn from a uniform distribution between $0$ and $1$ at the initialization of the landscape. The higher the value of $K$, the more the total payoff changes by a flip of a single element and the more rugged the landscape. We normalize the total payoff by the maximum payoff on a landscape ($P_{\text{norm}} = P/P_{\text{max}}$) and raise its value to the power of $8$ ($(P_{\text{norm}})^8)$), following past studies \cite{siggelkow2005speed,lazer2007network, barkoczi2016social}. For ease of explanation, we scale the payoff by a factor of $c_{\text{payoff}} = 100$ to normalize the possible payoff from $0$ to $100$. 

For social learning, we generate networks and let $n = 100$ agents receive the social information from their neighbors. For the fully connected network, every agent is connected to every other agent, fixing the degree of every node to $n - 1$. For the max mean clustering (MaxMC) network, we adopt the network structure from \cite{barkoczi2016social} where the network with a fixed degree of $19$ is iteratively rewired to maximize the mean clustering coefficient of the network.

\subsection{Baseline simulation procedure}
We assign uniformly random binary vectors to a group of $n$ individuals, and they perform the gathering and adoption process at each time step. First, they apply the given SLS by collecting information from randomly sampled $s$ agents among their connected neighbors. The established social option could be the solution of the agent with the highest payoff (best imitator, BI), the most frequent solution in the sample (conformist, CF), or just any random solution in the sample (random imitator, RI). Second, they adopt the social option if its payoff is greater than the current one, or otherwise, they perform individual (asocial) learning instead (and adopt it if its payoff is greater than the current one). In the case of CF, we also perform individual learning when all of the solutions are equally frequent. The pure individualist (PI) only performs individual learning and does not engage in any form of social learning.

There are several options for asocial learning, which is expressed with a hyphenated abbreviation: models with ``-I" (individual) perform exploration by flipping a randomly selected single bit from the current solution and adopting it if the payoff becomes higher; models with``-P" (probabilistic) assign an independent probability ($1/N$) for each dimension to be flipped (hence multiple bits can be flipped in a single step); and models with ``-R" (random) sample the asocial option completely randomly, without regard to the current state. Pure model names (BI, CF, RI) indicate that the model does not perform individual learning.

We repeat this procedure for $L=200$ steps and record the statistics. The test results, including the neural SLSs, are averaged across $100$ randomly initialized repetitions from $50$ different landscapes, hence a total of $5,000$ repetitions per SLS.

\subsection{Agent architecture and training method}
For reinforcement learning, we use a actor-critic algorithm with importance sampling and a generalized advantage estimator, which its implementation is identical to a standard proximal policy optimization (PPO) \cite{schulman2017proximal} method without any clipping on its policy. The discrete stochastic actor with categorical probability distribution is trained to receive the same information as the baseline SLS and yield a new solution (binary vector of length $N$) for each agent, while the critic aims to approximate the value function of the given state. We employ a general advantage estimator \cite{schulman2015high} for the advantage function with a discounting factor $\gamma=0.98$ and $\lambda=0.95$. When the goal of the neural SLS is to maximize the area under the average payoff curve, we provide the reward as a payoff of the produced solution at each step. When the goal is to maximize the final payoff at the final step, we provide the reward as a final payoff scaled by trajectory length $L$ at the final step.

Generally, there is no priority between sampled neighbors in SLSs, but the current learner should be able to discriminate itself from its neighbors. To handle this problem, we employ the set transformer \cite{lee2019set} as an architecture for the actor to guarantee permutation invariance and attach a binary variable to each agent's solution vector as a self-indicator ($1$ for the current agent and $0$ for sampled neighbors). To provide the exact ranking and frequency, we calculate (1) the competition ranking of scores among neighbors and itself ($1+N$ agents), and (2) the frequency of each state among its neighbors ($N$). We normalize these two features to let the terms have a scale of $0$ to $1$. The input is formed as a tensor of shape $B \times (S+1) \times (N+1+I)$, where $B$ is the batch size, $S$ is the number of neighbors, $N$ is a dimension of the given NK landscape, and $I$ is the number of features we provide. Here, $N+1+I$ indicates that the fitness of each solution ($1$) and additional information ($I$) are provided to the model.  After receiving this input, the actor yields a tensor of shape $B \times 2 \times N$, which represents the logit of the probability of $0$ and $1$ for each dimension. The final output is then sampled from the normalized logits and compared with the current solution. The one with a higher payoff becomes the solution of the agents for the next time step. This procedure is repeated for an episode length of $L=200$ steps.

We trained our model for $10,000$ epochs with early stopping, which took around $3$ to $4$ days using five Titan V GPUs in parallel computation. All of the experiments used the $tanh$ activation function and Adam optimizer \cite{kingma2014adam} with a learning rate of the actor of $1.0\times 10^{-5}$, learning rate of the critic of $3.0\times 10^{-5}$, and entropy coefficient of $0.0003$, without any further scheduling. In each epoch, every $100$th iteration sampled $1,000$ data from the replay buffer for computing losses for the actor and critic and performing a gradient update. All of the code is implemented in \texttt{PyTorch}.

\subsection{Strategy visualization}

For the BI test, we first obtain the model output probabilities for a test template with all possible integer pairs of payoff $(p_0, p_1, p_2, p_3)$, where $0 \leq p_0 \leq p_{\text{max}}$ and $0 \leq p_3 \leq p_2 \leq p_1 \leq p_{\text{max}}$. We can set this inequality without loss of generality and reduce the effective number of payoff triplets (for each $p_0$) from $(p_{\text{max}}+1)^3$ to $(p_{\text{max+1}})(p_{\text{max}}+2)(p_{\text{max}}+3)/6$ (which is $176,851$ for $p_{\text{max}} = c_{\text{payoff}} = 100$) because the SLS (and our model) is invariant to the neighbor permutation. The model output, corresponding to the probability of producing $1$, is plotted in the form of a 2D output diagram.

For the 3D voxel plot, we calculate the normalized Euclidean distances from (1) the self solution $\textbf{x}_{\text{self}}$, $d_{\text{self}} = \sqrt{\sum_i^N(x_i-x_{\text{self}, i})^2/N}$, (2) the solution with the second highest payoff (third row of the test template) $\textbf{x}_{\text{second}}$, $d_{\text{second}} = \sqrt{\sum_i^N(x_i-x_{\text{second}, i})^2/N}$, and (3) the solution with the highest payoff (second row of the test template) $\textbf{x}_{\text{best}}$, $d_{\text{best}} = \sqrt{\sum_i^N(x_i-x_{\text{best}, i})^2/N}$ to each model output $\textbf{x}$. Finally, we set $r = 1 - d_{\text{self}}$, $g = 1 - d_{\text{second}}$, $b = 1 - d_{\text{best}}$, and opacity $a = 0.3(1-\min({d_{\text{self}}, d_{\text{second}}, d_{\text{best}}}))^2$ for visualization. These values become the color code of the corresponding voxel's face, $(r, g, b, a)$.

For the CF test, we use a different test template having one neighbor with a strictly high payoff $p_1$ and two neighbors with the same solutions and lower payoff, $p_2 = p_3 < p_1$. Considering the permutation invariant, the effective number of payoff triplets (for each $p_0$) for this condition is $(p_{\text{max}})(p_{\text{max}}+1)/2$ (which is $5050$ for $p_{\text{max}} = 100$); the 2D output diagram depicts the model output for these inputs.

\section{Learning scheme with fixed environment}

Here, we present the result when instead of providing random NK landscapes at every epoch, only $1$ and $10$ fixed NK landscapes are given to the agent during the training (Fig. S1A). We find that in the case of a single landscape, the model performance quickly converges to the maximum value, $100$. This optimal performance is obtained from finding \textit{the best} solution ($[1, 1, 0, 0, 0, 1, 0, 1, 0, 1, 1, 0, 1, 0, 1]$ in this case) and precisely producing this solution regardless of the input (Fig. S1B). This result demonstrates the effectiveness of reinforcement learning as a meta-heuristic, but the agent failed to achieve a general sense of social learning and choose to memorize the answer and stick with it, which is intuitively the best strategy in this particular case. 

Interestingly, we can observe that the agent manages to achieve great performance (over $80$) when $10$ different landscapes are given (Fig. S1A). In this case, we employ $10$ GPU and parallel computing to train the agent. Since the memorization of a single solution would not yield a high average payoff for $10$ different landscapes, we check the model output with the Best-Imitator test (Fig. S1B) The model output seems like it tries to copy the third-best solution, but we perform the same test with different template and find that the model output looks similar regardless of the test template. Note that in this case, many dimensions of the output probabilities have neither $0$ nor $1$, but intermediate values. This implies that the model did not attempt to learn from others or memorize a single solution, but somehow found the probabilistic solution that can achieve good performance for all $10$ environments when iteratively applied to all agents. We test this hypothesis by evaluating the model performance with randomly generated NK landscapes. As we expected, our model shows a good mean payoff for its training landscapes but fails to achieve any meaningful performance for random landscapes. One notable thing is that the increase of mean payoff is not instant; our model's probabilistic solution needs to be iteratively applied in order to achieve its final solution. Although the existence of such a solution is intriguing on its own, the model failed to acquire any social learning skills in this environment. These results imply that the provision of enough different landscapes is vital for motivating agents to learn social skills rather than optimized to fixed environments.

\begin{figure}
  \centering
  \includegraphics[width=1\linewidth]{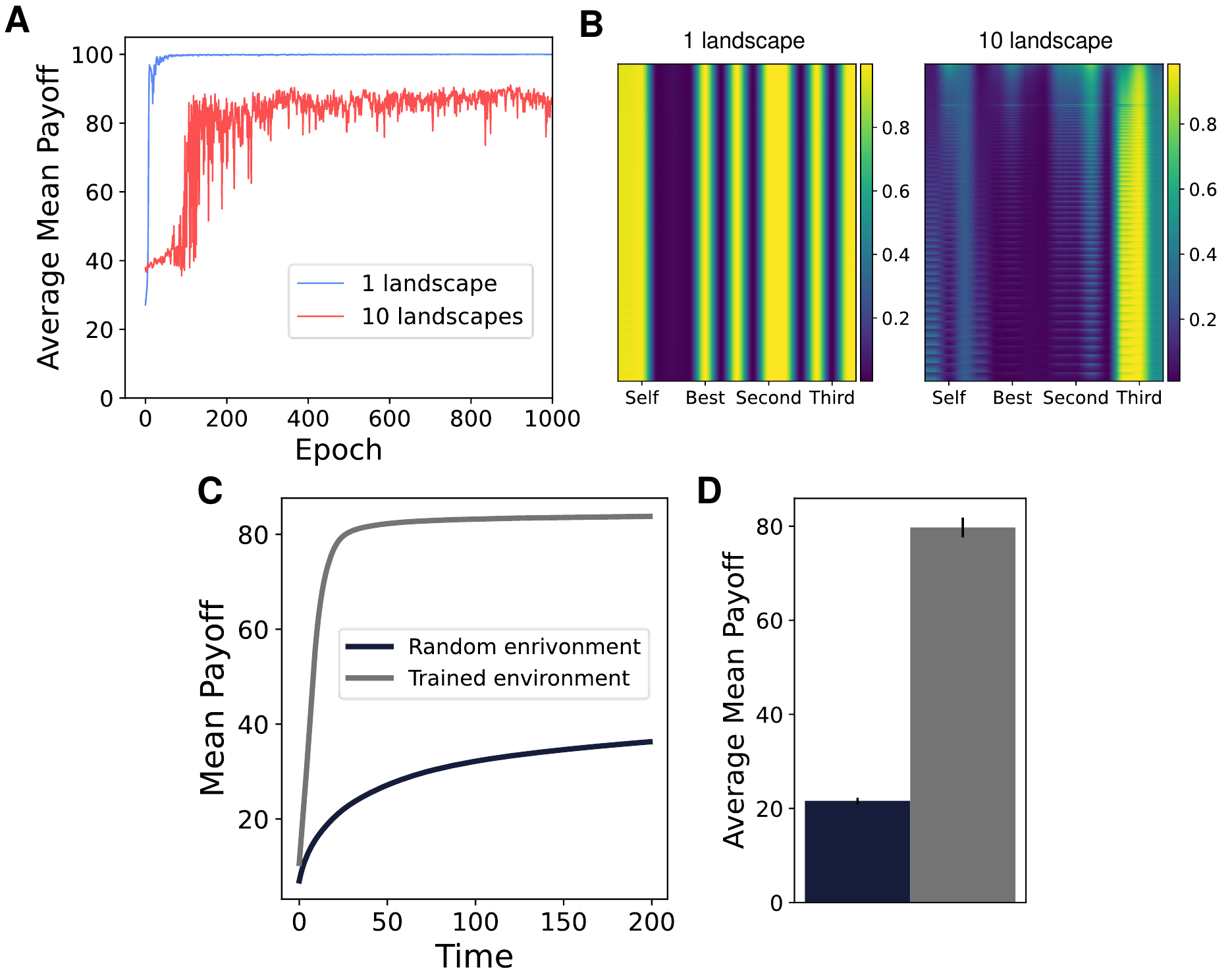}
  \caption{(A) Average mean payoff when $1$ (blue) and $10$ (red) fixed landscapes are given during the training, instead of randomly initialized landscape at each epoch. (B) Best-imitator test result of the final model from $1$ (left) and $10$ (right) landscapes. (C) Mean payoff and (D) average mean payoff of the final model from $10$ landscapes, evaluated with $10$ random landscapes (black) and $10$ training landscapes (gray).}\label{fig:s1}
\end{figure}

\section{Learning scheme with group-averaged reward}

Here, we present the result when instead of individual payoff, the group-averaged payoff is given as a reward at each timestep during the training. We find that the individual has failed to learn any form of social learning when the group-averaged reward is given (Fig. S1). We speculate that in order to tackle this problem from a group-focused viewpoint, one might need to specialize in the architecture and loss function for controlling the action of the entire group, such as a centralized controller. Note that in that case, the dimensionality of the group action would be enormous ($15\times 100 = 1500$ in our default case) so another form of bypass or remedy would be needed to reduce it effectively.

\begin{figure}
  \centering
  \includegraphics[width=1\linewidth]{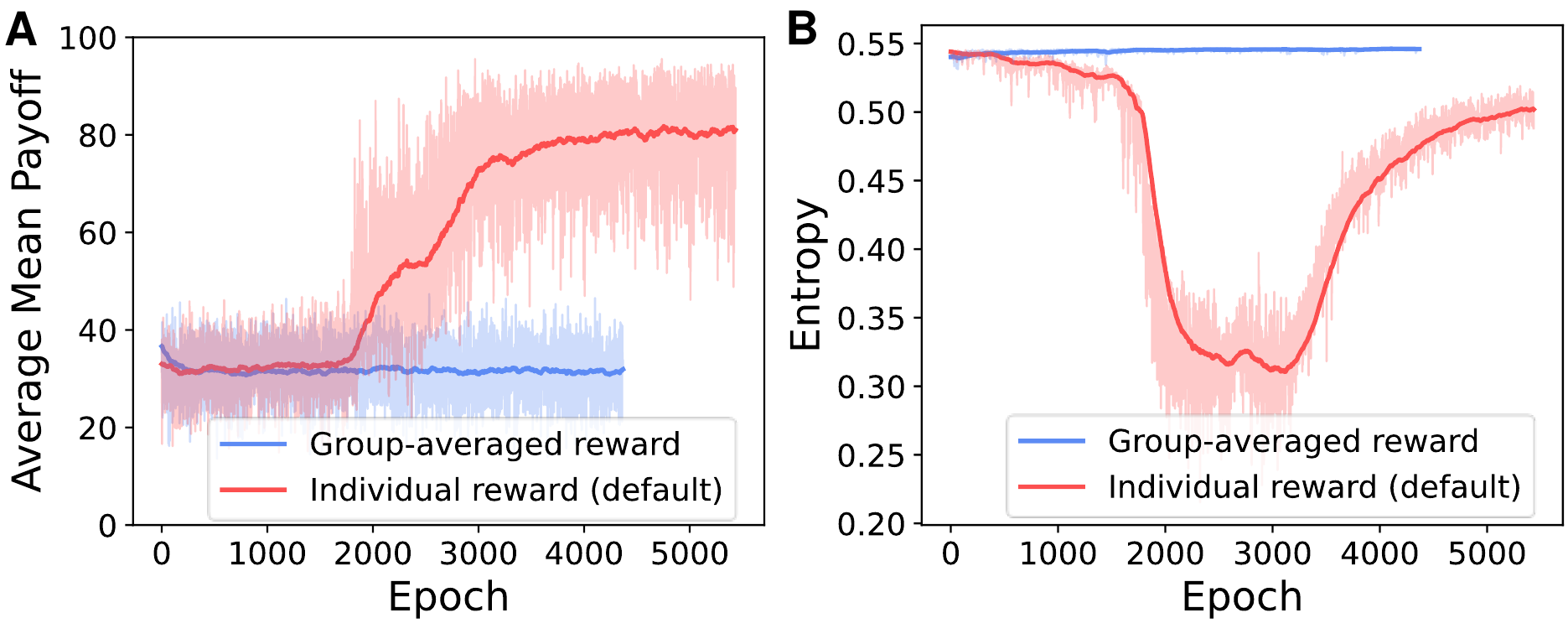}
  \caption{(A) Average mean payoff and (B) entropy of two model output, the model with group-averaged reward and the model with individual reward (default). The bold line shows an exponential moving average with a smoothing factor of $0.99$.}\label{fig:s2}
\end{figure}

\section{Training results when the frequency feature is not provided}

Here, we test the alternative provision of information by removing the frequency feature from the input. In the default settings, we provide $5$ different information of itself and $3$ neighbors to each agent without any structural information: binary solution vector of dimension $N$, payoff, self-indicator, ranking (including itself), and solution frequency (excepting itself). We name this setting as \textbf{PIRF} (\textbf{P}ayoff, \textbf{I}ndicator, \textbf{R}anking, \textbf{F}requency). By removing the solution frequency feature, the setting is then called \textbf{PIR} setting.

We plot the mean average payoff from five trials with different seeds for each setting (Fig. S3). We can observe that the default setting, PIRF advances the timing of the realization of copying compared to the PIR setting. As we explain in the main manuscript, we speculate that the additional information which can lead to performance improvement might boost the learning process by facilitating the acquisition of the concept of copying.

\begin{figure}
  \centering
  \includegraphics[width=0.7\linewidth]{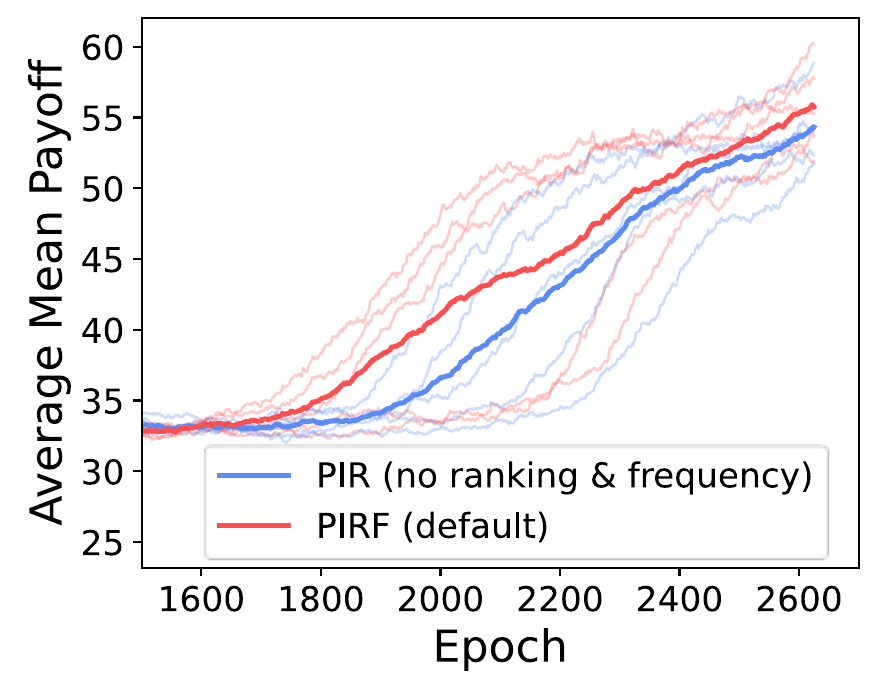}
  \caption{(A) Exponential moving average of average mean payoff from five trials of PIR and PIRF (default) settings (with a smoothing factor of $0.99$). The bold line shows an average of five trials.}\label{fig:s3}
\end{figure}

\section{Training results when the payoff ranking is not provided}

Here, we test the alternative provision of information by removing payoff ranking (and frequency feature) from the input, which is \textbf{PI} setting. We find that the final model of PI and PIRF is qualitatively similar (not shown), but the PI setting takes much longer training epochs to reach its final model (Fig. S4). We speculate that this slow convergence is because our model tries to learn a total order of all $4$ continuous payoff, which could be quite challenging without tailored architecture and loss function \cite{burges2005learning}. In this case, since we normalize payoff into $[0, 1]$ range, the task would be slightly easier and the model eventually succeeds to learn a well-performing ranking function.

\begin{figure}
  \centering
  \includegraphics[width=1\linewidth]{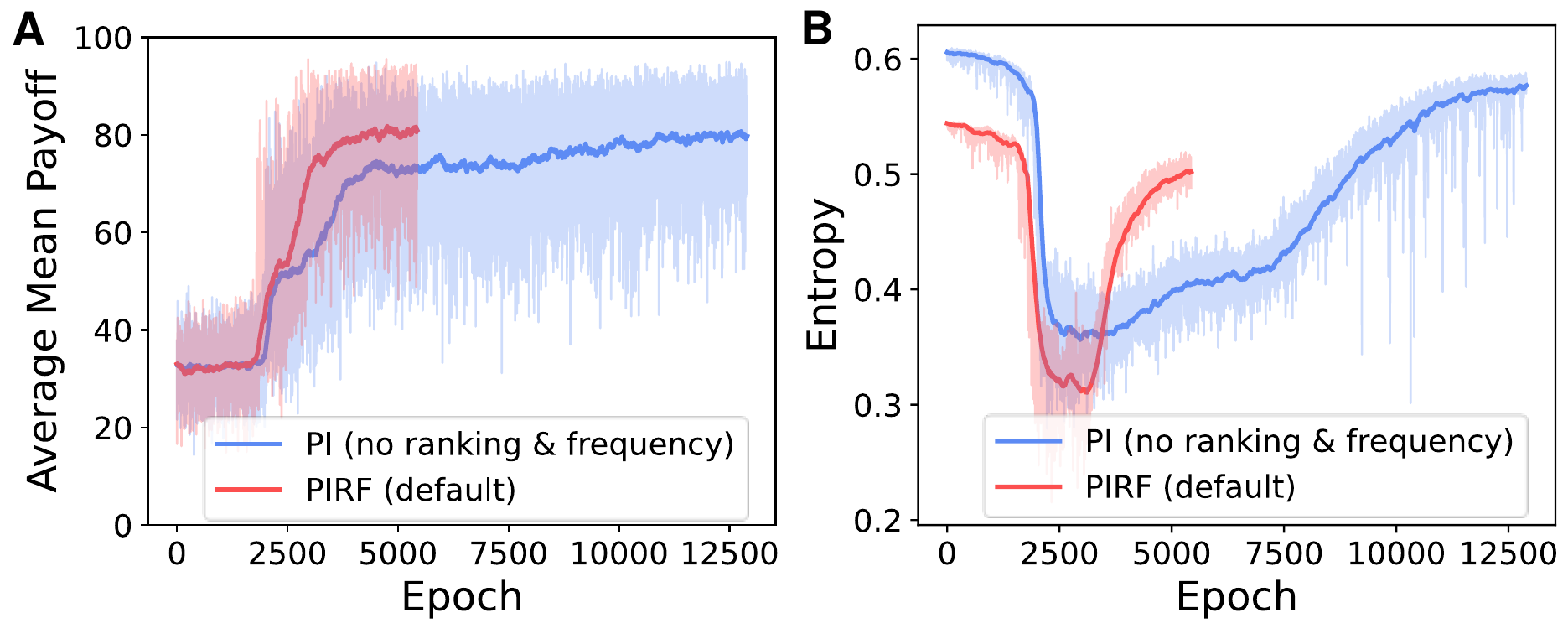}
  \caption{(A) Average mean payoff and (B) entropy of two model output, PI and PIRF (default). The bold line shows an exponential moving average with a smoothing factor of $0.99$.}\label{fig:s4}
\end{figure}

\section{Training results with more rugged landscape with $K=11$}

Here, we present the result from a more rugged environment compared to the default $K=7$ landscape, $K=11$. We find that due to its extreme ruggedness, the agent struggles to realize the very first step of social learning; the concept of copying (Fig. S5A, B). We trained the model for nearly $4,500$ epochs and the model stayed at the initial random strategy. We expect that increasing training epochs will eventually lead to a realization of copying (since we find that the model with a less rugged landscape tends to realize the concept of copying earlier), there are ways to boost this initial stage of learning.

When training the agent with a complex task, curriculum learning \cite{narvekar2020curriculum} helps the training by scheduling the level of difficulty from small and easy tasks to large and difficult tasks, successively and gradually. We can apply this technique in our problem settings by initially providing a less rugged landscape and then substituting it with a more rugged landscape afterward. We also plot the result with a curriculum learning scheme from three different scheduling; starting from $K=3$ landscape, each scheduler then changed the environment to $K=11$ landscape at $1,000$ (when the agent realizes the concept of copying), $2,500$ (when the agent learns to imitate the best), and $5,500$ epochs (the final model), respectively (Fig.S5A and B). In all three cases, the model successfully learns social learning without any problem, implying that the realization of copying is the sole and hardest barrier to pass. We show that the final model from scheduling of $2500$ epoch shows nearly similar performance compared to the BI-R model (Fig. S5C and D)

\begin{figure}
  \centering
  \includegraphics[width=1\linewidth]{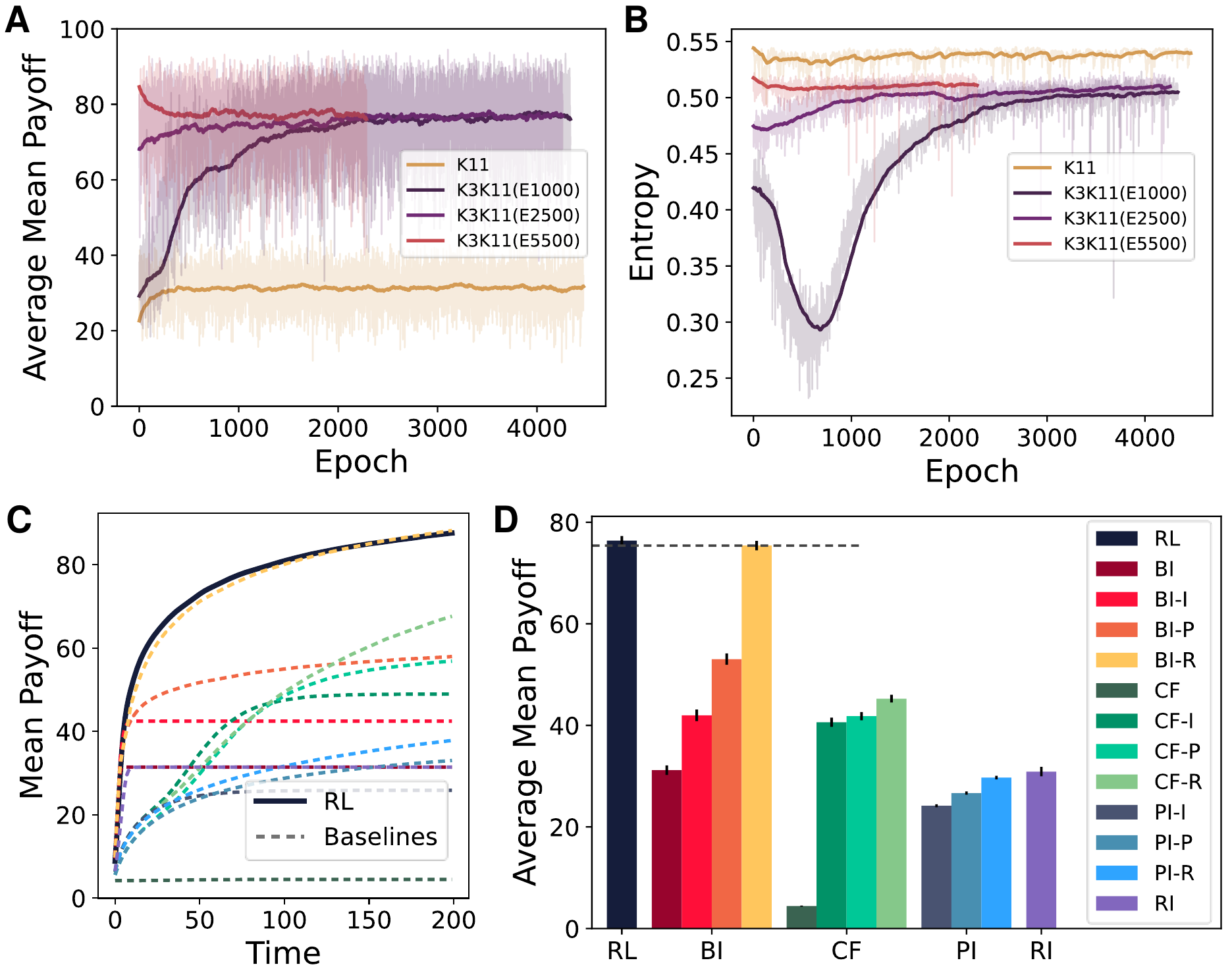}
  \caption{(A) Average mean payoff and (B) entropy of four model output, $K=11$ without any scheduling ($K11$), and with three different scheduling scheme which changes the landscape from $K=3$ to $K=11$ at epoch $1,000$ ($K3K11 (E1000)$), $2,500$ ($K3K11 (E2500)$), and $5,500$ ($K3K11 (E5500)$). The bold line shows an exponential moving average with a smoothing factor of $0.99$. (C) Mean payoff and (D) Average mean payoff over time of SLS from the final model from $K3K11 (E2500)$ (RL) and various baseline SLSs. Error bars show $\pm$5 standard error of the mean.}\label{fig:s5}
\end{figure}

\section{Verification with real social networks}

We adopted a network dataset from \cite{ghasemian2020stacking}, which contains $124$ social network structures among a total of $550$ networks. Since the conformist baseline needs at least $3$ neighbors to perform its SLS, we applied $k$-core decomposition to all $124$ networks with $k=3$ and check whether removed nodes are less than $5\%$ from its original network and the decomposed network is still connected. After the decomposition, a total of $88$ networks passed the criteria and we filtered networks with more than $500$ nodes. As a result, $53$ networks satisfied all conditions, and their node numbers range from $39$ to $478$ after the decomposition. We perform the same procedure as the default settings with all $53$ networks. The results are averaged across $20$ repetitions from $5$ different landscapes, hence a total of $100$ repetitions per network per SLS.

We find that similar to the other results, BI-R showed the best performance among the baselines and our model exceeds its performance, but by a small margin in this case (Fig. S6). The higher error bar is due to a smaller number of trials. Note that in this case, we do not individually train the model for each of the $53$ networks, but the default model (environment of complete network with $100$ agents) is used to test all of the results. In the main manuscript, we show that the characteristic such as the level of copying can be different by training environment, and some environment prefers a higher level of copying while other environment does not. We guess that in this case, by averaging all results from $53$ networks, many of the advantages of our model from delicate balancing might be canceled out and result in a small margin compared to the full-copying model (BI-R). We present this result to demonstrate that the final model of our framework is still powerful for $(15,7)$ settings in various forms and sizes of real social networks.

\begin{figure}
  \centering
  \includegraphics[width=1\linewidth]{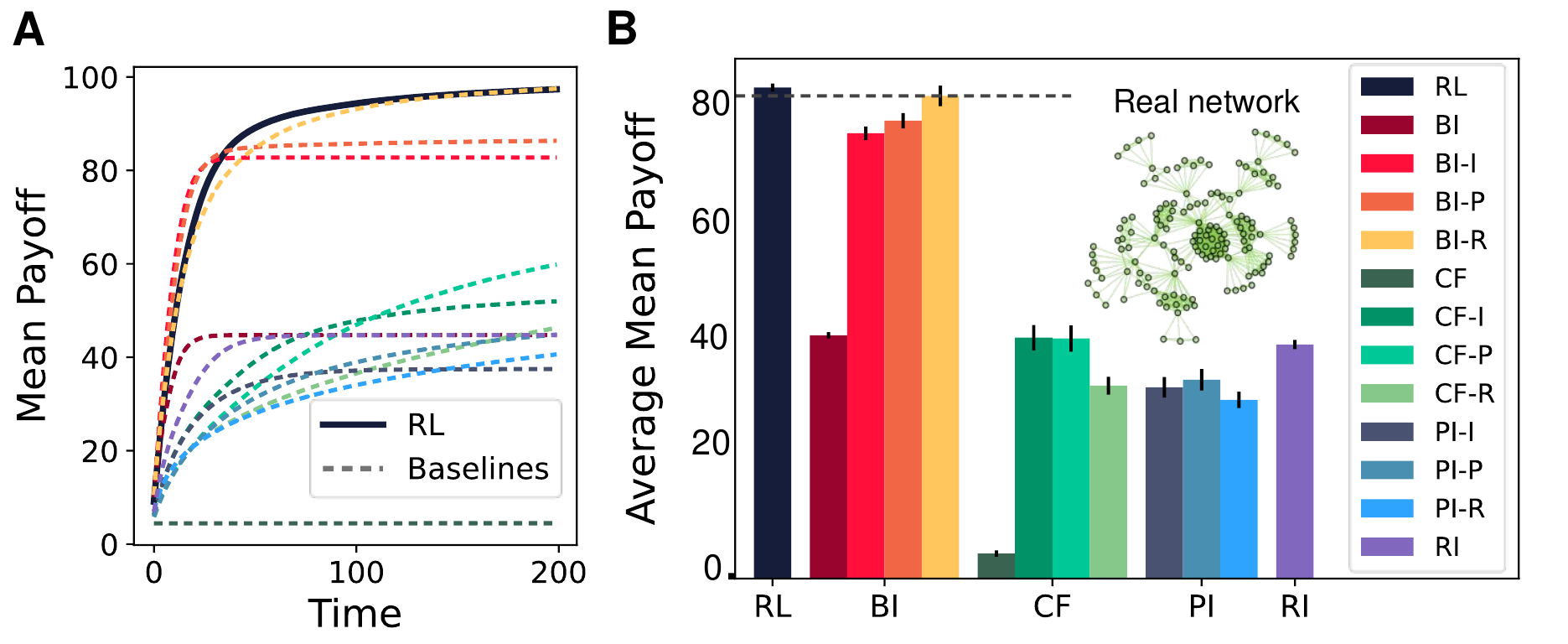}
  \caption{(A) Mean payoff and (B) Average mean payoff over time of SLS from reinforcement learning (RL) and various baseline SLSs. Error bars show $\pm$5 standard error of the mean.}\label{fig:s6}
\end{figure}

\section{Movie S1: Strategy transition throughout the training with the default setting}
Here, the movie shows the transition of the 3D strategy diagram, which spans nearly the entire training sequence (the movie ends at $5,400$ epochs where the final model is from $5,700$ epochs of training.) We can observe that various short-term strategical notions appear during the training (Fig. S7). For instance, during the transition between stage 2 and 3 (epoch $2,700$), the agent starts to copy the second-best strategy in some cases, depicted as green voxels. Considering that the agent during this transition learns a concept of 'copying based on payoff', copying the second-best solution can be regarded as a plausible intermediate strategy to passing by. Also, during the transition between stage 3 and 4 (epoch $3,800$), the agent adopts a different form of individual learning, depicted as red, opaque voxels, before it finally settles down to a total random flipping. Since the red color indicates high similarity with the current solution of the learner itself, this strategy is similar to the ``-P'' (probabilistic) individual learning in the baseline SLS where the agent keeps its original solution but allows some flipping. The agent eventually abandons such form of individual learning at the end of stage 4, since the ``-I'' individual learning shows superior performance compared to the ``-P'' individual learning when combined with BI strategy, as shown in Fig. 3B in the main manuscript. We speculate that the BI strategy goes well with highly exploratory individual learning because 'copying the best' strategy often causes early convergence due to its diversity-reducing nature, and exploration from the individual learning could complement such demerit. These intermediate strategical notions are indeed interesting observations by themselves, and also clearly demonstrate the vast scope of SLSs that our model can express.

\begin{figure}
  \centering
  \includegraphics[width=1\linewidth]{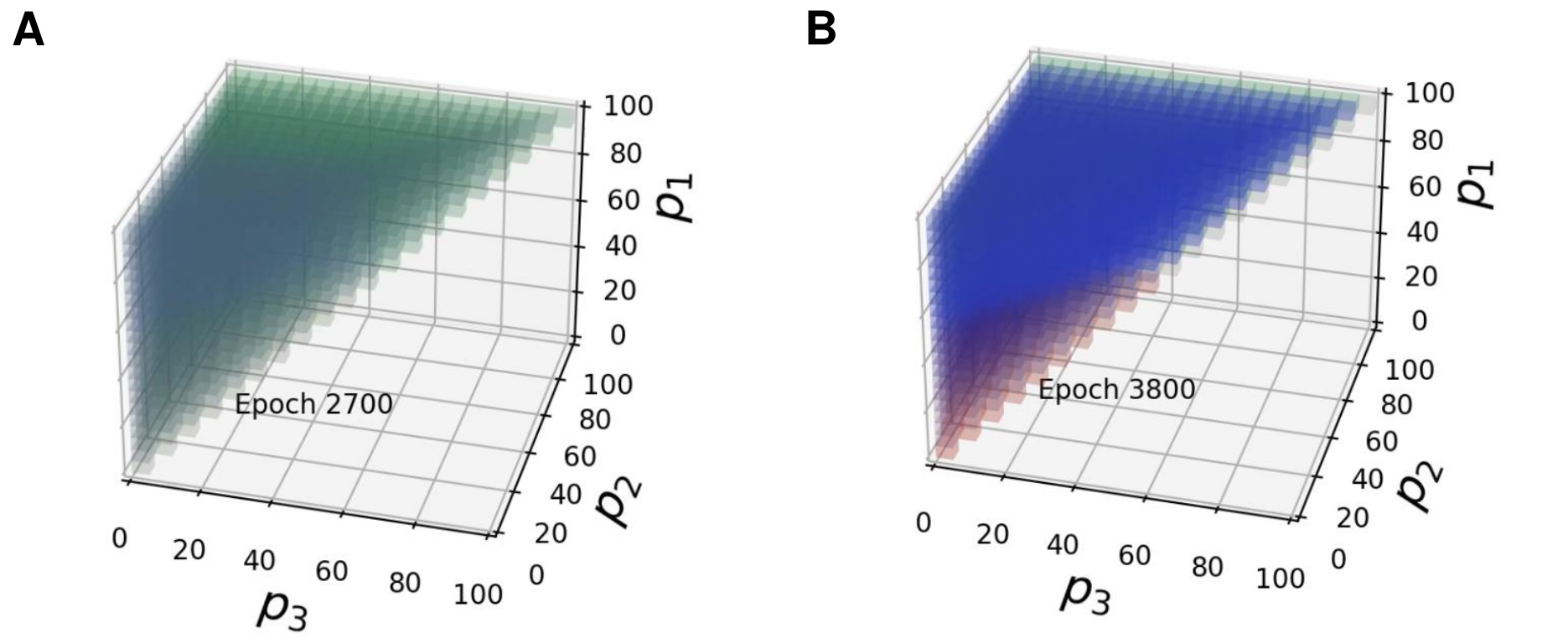}
  \caption{3D strategy diagrams from the model with (A) $2,700$ epochs and (B) $3,800$ epochs of training, which are the captures from the supplementary movie S1. The model is trained on a fully connected network of $100$ agents with NK$(15, 7)$ environments.}\label{fig:s7}
\end{figure}


\end{document}